\documentclass[final,OA]{AAAI-Std}

\def\artid{0001}%

\usepackage{times}
\usepackage{soul}
\usepackage{url}
\usepackage[utf8]{inputenc}
\usepackage{graphicx}
\usepackage{amsmath}
\usepackage{amsthm}
\usepackage{amsfonts}
\usepackage{booktabs}
\usepackage{algorithm}
\usepackage{algorithmic}
\usepackage{array,multirow,graphicx}
\usepackage{wasysym}
\usepackage{wrapfig}
\usepackage{csquotes}
\usepackage{xcolor}
\usepackage{color, colortbl}
\usepackage{tikz}
\usepackage{longtable,lscape}

\definecolor{LightCyan}{rgb}{0.88,1,1}
\definecolor{Gray}{gray}{0.9}
\newcommand{\pie}{%
/
}
\newcommand{\piee}{%
\Circle
}
\newcommand{\pieh}{%
\LEFTcircle
}
\newcommand{\pief}{%
\CIRCLE
}
\long\def\c#1{{\footnotesize{\fontfamily{pcr}\selectfont{#1}}}}
\long\def\cm#1{{\checkmark}}

\begin{document}

\title{A Survey of Knowledge-based Sequential Decision Making under Uncertainty}

\author{Shiqi Zhang}
\author{Mohan Sridharan}

\address{Shiqi Zhang is an Assistant Professor of Computer Science at the State University of New York (SUNY) at Binghamton, USA. Mohan Sridharan is a Reader in Cognitive Robot Systems in the School of Computer Science at the University of Birmingham, UK. }

\abstract[Abstract]{Reasoning with declarative knowledge (RDK) and sequential decision-making (SDM) are two key research areas in artificial intelligence. RDK methods reason with declarative domain knowledge, including commonsense knowledge, that is either provided a priori or acquired over time, while SDM methods (probabilistic planning and reinforcement learning) seek to compute action policies that maximize the expected cumulative utility over a time horizon; both classes of methods reason in the presence of uncertainty. Despite the rich literature in these two areas, researchers have not fully explored their complementary strengths. In this paper, we survey algorithms that leverage RDK methods while making sequential decisions under uncertainty. We discuss significant developments, open problems, and directions for future work. }

\keywords{diffusion paths, drift estimation, multiclass
classification, plug-in estimators}

\maketitle

\section{Introduction}
\label{sec:intro}
Agents operating in complex domains often have to execute a sequence of actions to complete complex tasks. These domains are characterized by non-deterministic action outcomes and partial observability, with sensing, reasoning, and actuation associated with varying levels of uncertainty. For instance, state of the art manipulation and grasping algorithms still cannot guarantee that a robot will grasp a desired object (say a coffee mug). 
In this paper, we use sequential decision making~(\textbf{SDM}) to refer to algorithms that enable agents in such domains to compute action policies that map the current state (or the agent's estimate of it) to an action. More specifically, we consider SDM methods that model uncertainty probabilistically, i.e., \textbf{probabilistic planning} and \textbf{reinforcement learning} methods that enable the agents to choose actions toward maximizing long-term utilities. 

SDM methods, by themselves, find it difficult to make best use of \emph{commonsense} knowledge that is often available in any given domain. This knowledge includes \textit{default} statements that hold in all but a few exceptional circumstances, e.g., ``books are usually in the library but cookbooks are in the kitchen'', but may not necessarily be natural or easy to represent quantitatively (e.g., probabilistically). It also includes information about domain objects and their attributes, agent attributes and actions, and rules governing domain dynamics. In this paper, we use \textbf{declarative knowledge} to refer to such knowledge represented as relational statements. Many methods have been developed for reasoning with declarative knowledge~(\textbf{RDK}), often using logics. These methods, by themselves, do not support probabilistic models of uncertainty toward achieving long-term goals, whereas a lot of information available to agents in dynamic domains is represented quantitatively to model the associated uncertainty.

\begin{figure*}[tb] 
\centering \vspace{.5em}
\includegraphics[width=0.45\textwidth]{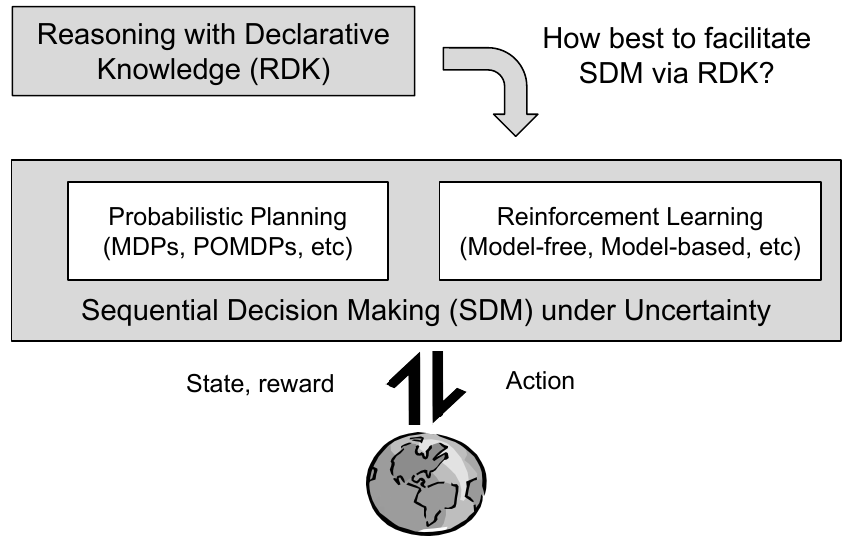}
\vspace{-0.5em}
\caption{An overview of this survey: \emph{reasoning with declarative knowledge} (RDK) for \emph{sequential decision making} (SDM). } 
\label{fig:overview}
\end{figure*}

For many decades, the development of RDK and SDM methods occurred in different communities that did not have a close interaction with each other. Sophisticated algorithms have been developed, more so in the last couple of decades, to combine the principles of RDK and SDM. However, even these developments have occurred in different communities, e.g., statistical relational AI, logic programming, reinforcement learning, and robotics. Also, these algorithms have not always considered the needs of agents in dynamic domains, e.g., reliability and computational efficiency while reasoning with incomplete knowledge. As a result, the complementary strengths of RDK and SDM methods have not been fully exploited. Also, figuring out how best to combine the principles of RDK and SDM remains an open grand challenge in AI, with connections to deep philosophical questions about the representation, manipulation/use, and acquisition of knowledge in humans and machines, and about the broader impacts of such methods. This survey paper seeks to stimulate cross-pollination of ideas between the communities working on different aspects of this grand challenge, by highlighting the key achievements and open problems. To achieve this objective while keeping the list of related papers manageable, we limit our scope to algorithms that use RDK to facilitate SDM, and focus on the following question:
\begin{displayquote}
\begin{center}
\emph{How best to reason with declarative knowledge for sequential decision making under uncertainty? }
\end{center}
\end{displayquote}
We also limit our attention to algorithms developed for an agent making sequential decisions under uncertainty in dynamic domains. Furthermore, to explain the key concepts, we often draw on our expertise in developing such methods for robots. Figure~\ref{fig:overview} provides an overview of the survey's theme\footnote{This survey is based on a tutorial, titled ``\emph{Knowledge-based Sequential Decision-Making under Uncertainty}'', presented by the authors at the AAAI Conference in 2019.}. We begin by describing some key concepts related to RDK and SDM systems (Section~\ref{sec:background}), followed by the factors we use to characterize the RDK-for-SDM systems (Section~\ref{sec:factors}). We then describe some representative RDK-for-SDM systems (Section~\ref{sec:rdkforsdm}) and discuss open problems in the design and use of such systems (Section~\ref{sec:challenges-opportunities}).

\section{Background}
\label{sec:background}
We begin by briefly introducing key concepts related to the RDK and SDM methods that we consider in this paper.

\subsection{Reasoning with Declarative Knowledge}
\label{sec:background-rdk}
We consider a representation of commonsense knowledge in the form of statements describing relations between domain objects, domain attributes, actions, and axioms (i.e., rules). Historically, declarative paradigms based on logics have been used to represent and reason with such knowledge. This knowledge can also be represented quantitatively, e.g., using probabilities, but this is not always meaningful, especially in the context of statements of default knowledge such as ``people typically drink a hot beverage in the morning'' and ``office doors usually closed over weekends''. In this survey, \emph{any mention of RDK refers to the use of logics for representing and using such domain knowledge for inference, planning, and diagnostics}. Planning and diagnostics in the context of RDK refer to \textit{classical planning}, i.e., computing a sequence of actions to achieve any given goal, monitoring the execution of actions, and replanning if needed. This is different from probabilistic planning that computes and uses policies to choose actions in any given state or belief state (Section~\ref{sec:background-sdm}). 

Prolog was one of the first logic programming languages~\citep{colmerauer1996birth}, encoding domain knowledge using ``rules'' in terms of relations and axioms. Inferences are drawn by running a \emph{query} over the relations. 
An axiom in Prolog is of the form: \begin{quote}
\begin{footnotesize}
\begin{verbatim}
Head :- Body
\end{verbatim}
\end{footnotesize}
\end{quote}
and is read as ``Head is true if Body is true''. For instance, the following rule states that all birds can fly. 
\begin{quote}
\begin{footnotesize}
\begin{verbatim}
fly(B) :- bird(B)
\end{verbatim}
\end{footnotesize}
\end{quote}

Rules with empty bodies are called \emph{facts}. For instance, we can use ``\c{bird(tweety)}'' to state that tweety is a bird. Reasoning with this fact and the rule given above, we can infer that ``\c{fly(tweety)}'', i.e., tweety can fly. 
Research on RDK dates back to the 1950's, and has produced many knowledge representation and reasoning paradigms, such as First Order Logic, Lambda Calculus~\citep{barendregt1984lambda}, Web Ontology Language~\citep{mcguinness2004owl}, and LISP~\citep{mccarthy1978history}.

\paragraph{Incomplete Knowledge}
In most practical domains, it is infeasible to provide \emph{comprehensive} domain knowledge. As a consequence, reasoning with the incomplete knowledge can result in incorrect or sub-optimal outcomes. Many logics have been developed for reasoning with incomplete declarative knowledge. One representative example is Answer set programming (ASP), a declarative paradigm~\citep{gebser:aspbook12,gelfond2014knowledge}. ASP supports \emph{default negation} and \emph{epistemic disjunction} to provide non-monotonic logical reasoning, i.e., unlike classical logic, it allows an agent to revise previously held conclusions. An ASP program consists of a set of rules of the form: 
\begin{quote}
\begin{footnotesize}
\begin{verbatim}
a :- b, ..., c, not d, ..., not e. 
\end{verbatim}
\end{footnotesize}
\end{quote}
where \c{a...e} are called literals, and \c{not} represents default negation, i.e., \c{not d} implies that \c{d} is not believed to be true, which is different from saying that \c{d} is false.  Each literal can thus be true, false or unknown, and an agent associated with a program comprising such rules only believes things that it is forced to believe.

\paragraph{Action Knowledge}
Action languages are formal models of part of natural language used for describing transition diagrams, and many action languages have been developed and used in robotics and AI. This includes STRIPS~\citep{fikes1971strips}, PDDL~\citep{haslum2019introduction}, and those with a distributed representation such as $\mathcal{AL}_d$~\citep{Gelfond:ANCL13}. The following shows an example of using STRIPS to model an action \c{stack} whose preconditions require that the robot be holding object \c{X} and that object \c{Y} be clear. After executing this action, object \c{Y} is no longer clear and the robot is no longer holding \c{X}. 
\begin{quote}
\begin{footnotesize}
\begin{verbatim}
operator(stack(X,Y),
         Precond [holding(X),clear(Y)],
         Add [on(X,Y),clear(X)],
         Delete [holding(X),clear(Y)])
\end{verbatim}
\end{footnotesize}
\end{quote}


Given a goal, e.g., \c{on($b_1$,$b_2$)}, which requires block $b_1$ to be on $b_2$, the action language description, along with a description of the initial/current state, can be used for planning a sequence of actions that achieve this goal. 
Action languages and corresponding systems have been widely used for classical planning~\citep{ghallab2016automated}, aiming at computing action sequences toward accomplishing complex tasks that require more than one action. 

\paragraph{Hybrid Representations}
Logic-based knowledge representation paradigms typically support Prolog-style statements that are either true or false. By themselves, they do not support reasoning about quantitative measures of uncertainty, which is often necessary for the interactions with SDM paradigms. As a result, many RDK-for-SDM methods utilize hybrid knowledge representation paradigms that jointly support both logic-based and probabilistic representations of knowledge; they do so by associating probabilities with specific facts and/or rules. Over the years, many such paradigms have been developed; these include Markov Logic Network (MLN)~\citep{richardson:ML06}, Bayesian Logic~\citep{milch:bookchap07},  probabilistic first-order logic~\citep{halpern:book03}, PRISM~\citep{gorlin:TPLP12}, independent choice logic~\citep{poole:JLP00}, ProbLog~\citep{fierens:TPLP15,deraedt:ML15}, KBANN~\citep{towell1994knowledge}, and P-log, an extension of ASP~\citep{baral:TPLP09}. We will discuss some of these later in this paper.

\subsection{Sequential Decision Making}
\label{sec:background-sdm}
We consider two classes of SDM methods: probabilistic planning (\textbf{PP})~\citep{puterman2014markov} and reinforcement learning (\textbf{RL})~\citep{sutton2018reinforcement}, depending on the availability of world models. A common assumption in these methods is the first-order Markov property, i.e., the next state is assumed to be conditionally independent of all previous states given the current state. Also, actions are assumed to be non-deterministic, i.e., they do not always provide the expected outcomes, and the state is assumed to be fully or partially observable. Unlike classical planning (see Section~\ref{sec:background-rdk}), these methods compute and use a \textit{policy} that maps each possible (belief) state to an action to be executed in that (belief) state.

\paragraph{Probabilistic Planning}
If the state is fully observable, PP problems are often formulated as a Markov decision process (\textbf{MDP}) described by a four-tuple $\langle \mathcal{S}, \mathcal{A}, T, R \rangle$ whose elements define the set of states, set of actions, the probabilistic state transition function $T:\mathcal{S}\times \mathcal{A}\times \mathcal{S}\rightarrow [0,1]$, and the reward specification $R:\mathcal{S}\times \mathcal{A}\times\mathcal{S}'\rightarrow \Re$. Each state can be specified by assigning values to a (sub)set of domain attributes. The MDP is solved to maximize the expected cumulative reward over a time horizon, resulting in a \emph{policy}  $\pi:s\mapsto a$ that maps each state $s\in \mathcal{S}$ to an action $a\in\mathcal{A}$. Action execution corresponds to repeatedly invoking the policy and executing the corresponding action.

If the current world state is not fully observable, PP problems can be modeled as a partially observable MDP (\textbf{POMDP})~\citep{kaelbling1998planning} that is described by a six-tuple $\langle \mathcal{S}, \mathcal{A}, Z, T, O, R \rangle$, where $Z$ is a set of observations, and $O:\mathcal{S}\times \mathcal{A}\times Z\rightarrow [0,1]$ is the observation function; other elements are defined as in the case of MDPs. The agent maintains a \emph{belief state}, a probability distribution over the underlying states. It repeatedly executes actions, obtains observations, and revises the belief state through Bayesian updates:
\begin{align*}
\label{eqn:belief_update}
  b'(s') = \frac{O(s',a,o)\sum_{s\in \mathcal{S}}T(s,a,s')
    b(s)}{pr(o|a, b)}
\end{align*}
where $b$, $s$, $a$, and $o$ represent belief state, state, action, and observation respectively; and $pr(o|a,b)$ is a normalizer. The POMDP is also solved to maximize the expected cumulative reward over a time horizon; in this case, the output is a \emph{policy}  $\pi:b\mapsto a$ that maps beliefs to actions. 

\paragraph{Reinforcement Learning}
Agents frequently have to make sequential decisions with an incomplete model of domain dynamics (e.g., without $R$, $T$, or both), making it infeasible to use classical PP methods. Under such circumstances, RL algorithms can be used by the agent to explore the effects of executing different actions, learning a policy (mapping states to actions) that maximizes the expected cumulative reward as the agent tries to achieve a goal~\citep{sutton2018reinforcement}. The underlying formulation is that of an MDP or a formulation that reduces to an MDP under certain constraints. 

There are at least two broad classes of RL methods: \textbf{model-based} and \textbf{model-free}. Model-based RL methods enable an agent to learn a model of the domain, e.g., $R(s,a)$ and $T(s,a, s')$ in an MDP, from the experiences obtained by the agent by trying out different actions in different states. Once a model of the domain is learned, the agent can use PP methods to compute an action policy. Model-free RL methods, on the other hand, do not learn an explicit model of the domain; the policy is instead directly computed from the experiences gathered by the agent. The standard approach to incrementally update the value of each state is the Bellman equation:
$$v_{k+1}(s) = \sum_a \pi(a|s) \sum_{s',r} pr(s', r| s,a)[r+\gamma v_k(s')], \forall s\in \mathcal{S}$$
where $v(s)$ is the value of state $s$, and $\gamma$ is a discount factor. It is also possible to compute the values of state-action pairs, i.e., $Q(s, a)$, from which a policy can be computed.

Many algorithms have been developed for model-based and model-free RL; for more details, please see~\citep{sutton2018reinforcement}. More recent work has also explored the integration of deep neural networks (DNNs) with RL, e.g., to approximate the value function~\citep{mnih2015human}, and deep policy-based methods, e.g.,~\citep{schulman2015trust,schulman2017proximal}. This survey focuses on the interplay between SDM (including RL) and RDK methods; the properties of individual RL methods (or SDM methods) are out of scope.


\section{Characteristic Factors}
\label{sec:factors}
Before we discuss RDK-for-SDM algorithms and systems, we describe the factors that we use to characterize these systems. The first two factors are related to the representation of knowledge and uncertainty, and the next three factors are related to reasoning with this knowledge and the underlying assumptions about domain dynamics and observability. The final three factors are related to the acquisition of domain knowledge. Unless stated otherwise, the individual factors are orthogonal to each other, i.e., the choice of a particular value for one factor can (for the most part) be made independent of the choice of value for the other factors.

\begin{figure*}[tb] 
\centering 
\includegraphics[width=0.95\textwidth]{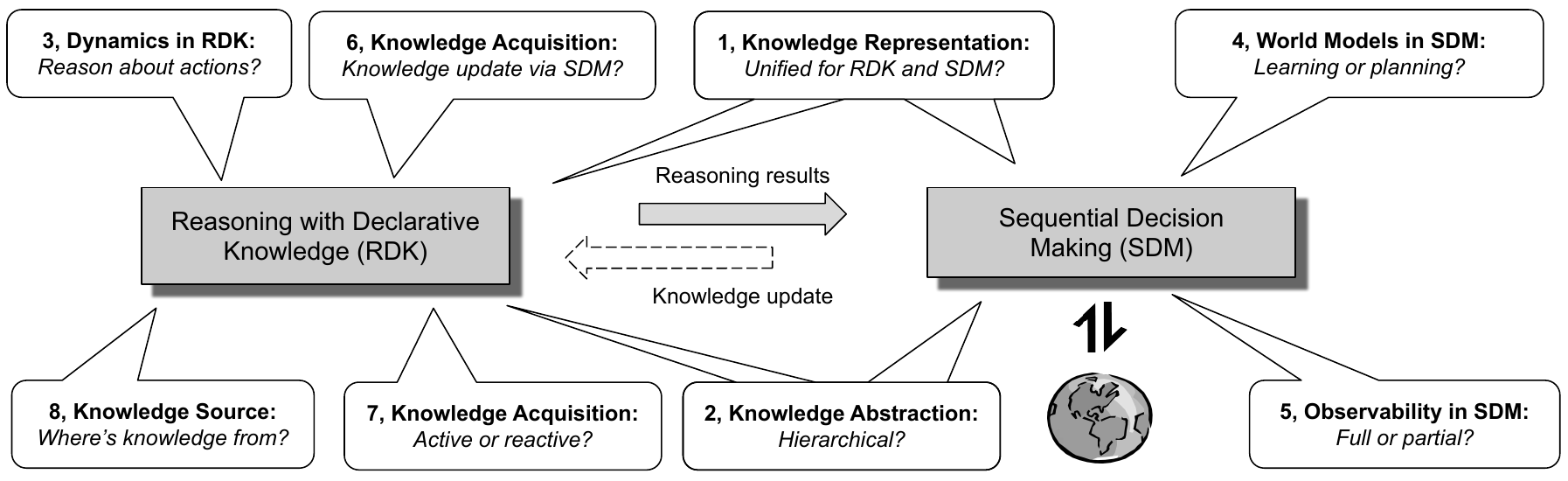}
\caption{Characteristic factors in the development of RDK-for-SDM methods. The individual factors are discussed in details in Section~\ref{sec:factors}, and are also used for the discussions of representative algorithms in Section~\ref{sec:rdkforsdm}. } 
\label{fig:choices}
\end{figure*}

\subsection{Representational Factors}
\label{sec:factors-represent}
We introduce two characteristic factors related to the representation of knowledge and uncertainty in RDK-for-SDM methods. The first factor is based on the relationships between the different descriptions of knowledge and uncertainty considered in these methods, and the second factor is based on the abstractions considered  in this representation.

\paragraph{Factor 1: Representation of Descriptions}
The \underline{\textit{first factor}} categorizes the methods that leverage RDK for SDM into two broad classes based on how they represent logical and probabilistic descriptions of knowledge and uncertainty.

Methods in the first group use a \textbf{unified representation} that is expressive enough to serve as the shared representation paradigm of both RDK and SDM. For instance, one can use a joint probabilistic-logical representation of knowledge and the associated uncertainty with probabilistic relational statements to describe both the beliefs of RDK, and the rules of SDM governing domain dynamics. These approaches provide significant expressive power, but manipulating such a representation (for reasoning or learning) imposes a significant computational burden. 

Methods in the second group use a \textbf{linked representation} to model the components of RDK and SDM. For instance, information closer to the sensorimotor level can be represented quantitatively to model the uncertainty in beliefs, and logics can be used for representing and reasoning with a more high-level representation of commonsense domain knowledge and uncertainty. 
These methods trade expressivity, correctness guarantees, or both for computational speed. 
For instance, to save computation, sometimes probabilistic statements with residual uncertainty are committed as true statements in the logical representation, potentially leading to incorrect inferences. 
Methods in this group can vary substantially based on if and how information and control are transferred between the different representations. For instance, many methods based on a linked representation switch between a logical representation and a probabilistic representation depending on the task to be performed; other methods address the challenging problem of establishing links between the corresponding transition diagrams to provide a \textit{tighter coupling}.

\paragraph{Factor 2: Knowledge Abstraction} 
RDK-for-SDM methods often reason about knowledge at different granularities.
Consider a mobile delivery robot. It can reason about rooms and cups to compute a high-level plan for preparing and delivering a beverage, and reason about geometric locations at finer granularities to grasp and manipulate a given cup. 

The \underline{\emph{second factor}} used to characterize these methods is the use of different \textbf{abstractions} within each component or in different components, e.g., a hierarchy of state-action spaces for SDM, or a combination of an abstract representation for logic-based task planning and a fine-resolution metric representation for probabilistic motion planning. The use of different abstractions makes it difficult to identify and use all the relevant knowledge to make decisions; note that this challenge is present in the ``linked representation'' methods discussed in the context of Factor 1.

Methods that explore different abstractions of knowledge often encode rich domain knowledge (including cognitive models) and perform RDK at an abstract level, using SDM for selecting and executing more primitive (but more precise) actions at a finer granularity.
Despite the variety of knowledge representation paradigms, we still maintain the constraint that we only consider algorithms with knowledge being represented declaratively.

\subsection{Reasoning Factors}
\label{sec:factors-reason}
Given a representation of knowledge, an agent needs to reason with this knowledge to achieve desired goals. Here, we are particularly interested in if and how knowledge of domain dynamics is used in the RDK and SDM components, and the effect of state representation on the choice of methods. These issues are captured by the following three factors. 

\paragraph{Factor 3: Dynamics in RDK}
RDK algorithms manipulate the underlying representations for different classes of tasks; in this paper, we consider inference, classical planning, and diagnostics. Among these tasks, inference requires the agent to draw conclusions based on its current beliefs and domain knowledge, while planning and diagnostics require RDK algorithms to reason about changes caused by actions executed over a period of time. 

We introduce the \underline{\emph{third factor}} to categorize the RDK-for-SDM methods into two groups based on whether the RDK component reasons about \textbf{actions and change}. 
Methods that perform inference based on a particular snapshot of the world are in one category, whereas methods for classical planning and diagnostics that require decisions to be made over a sequence of time steps are in the other category. These choices may be influenced by the specific application or how RDK is used for SDM methods. 
For instance, RDK-for-SDM methods that leverage domain knowledge to improve the exploration behaviors of RL agents require the ability to reason about actions and change.

\paragraph{Factor 4: World Models in SDM}
The \underline{\textit{fourth factor}} categorizes SDM methods, i.e., PP and RL methods, depending on the availability of world models. When world models are available, we can construct Dynamic Bayesian Networks (DBNs), and compute action policies. When world models are not available, the agent can interact with its environment and learn action policies through trial and error, and this can be formulated as an RL problem. Among the RL methods, model-based methods explicitly learn the domain dynamics (e.g., $T$ and $R$) and use PP methods to compute the action policy. Model-free RL methods, on the other hand, enable agents to directly use their experiences of executing different actions to compute the policies for mapping states to actions.

\paragraph{Factor 5: State or Belief State in SDM}
SDM methods involve an agent making decisions based on observing and estimating the state of the world. A key distinction here is whether this state is fully observable or partially observable, or equivalently, whether the observations are assumed to be complete and correct. The \underline{\textit{fifth factor}} categorizes the SDM methods based on whether they reason assuming full knowledge of state after action execution, or assume that the true state is unknown and reason with \textit{belief states}, e.g., probability distributions over the underlying states~\citep{kaelbling1998planning}, and implicit representations computed with neural networks~\citep{SDMIA15-Hausknecht}. Among the SDM formulations considered in this paper, MDPs map to the former category, whereas POMDPs map to the latter category. 

\medskip
Note that there are other distinguishing characteristics of reasoning systems (of RDK-for-SDM systems) that we do not explore in this paper. For instance, reasoning in such systems often includes a combination of active and reactive processes, e.g., actively planning and executing a sequence of actions to achieve a particular goal in the RDK component, and computing a probabilistic policy that is then used reactively for action selection in the SDM component.

\subsection{Knowledge Acquisition Factors}
\label{sec:factors-acquire}
Since comprehensive domain knowledge is often not available, RDK-for-SDM methods may include an approach for knowledge acquisition and revision. We introduce three characteristic factors related to how knowledge is acquired and the source of this knowledge.

\paragraph{Factor 6: Online vs. Offline Acquisition}
Our \underline{\emph{sixth factor}} categorizes methods based on whether they acquire knowledge \textbf{online} or \textbf{offline}. Methods in the first category interleave knowledge acquisition and task completion, with the agent revising its knowledge while performing the task based on the corresponding observations. In comparison, methods in the second category decouple knowledge acquisition and task completion, with the agent extracting knowledge from a batch of observations in a separate phase; this phase occurs either before or after the assigned task is completed. For the purposes of this survey, RDK-for-SDM methods that do not support knowledge acquisition are grouped in the ``offline'' category for this characteristic factor. 


\paragraph{Factor 7: Active vs. Reactive Acquisition}
RDK-for-SDM methods are categorized by our \underline{\textit{seventh factor}} based on whether they explicitly execute actions for acquiring knowledge. Some methods use an \textbf{active acquisition} approach, which has the agent explicitly plan and execute actions with the objective of acquiring previously unknown knowledge and revising existing knowledge. These actions take the form of exploring the outcomes of actions and extracting information from the observations, or soliciting input from humans. Active acquisition is often coupled with active or reactive reasoning, e.g., for computing exploration plans. Other RDK-for-SDM methods use a \textbf{reactive acquisition} approach in which knowledge acquisition is a secondary outcome. As the agent is executing actions to perform the target task(s), it ends up acquiring knowledge from the corresponding observations; this may, in turn, trigger active acquisition.

\paragraph{Factor 8: Knowledge Source}
The \underline{\textit{eighth factor}} categorizes RDK-for-SDM methods based on the source of the declarative domain knowledge. In some methods, this knowledge is obtained by direct \textbf{human encoding}, e.g., in the form of logical statements written by humans to represent facts and axioms. This is a common source of domain knowledge, especially that which is encoded initially. The knowledge can also be acquired through \textbf{agent interaction}, e.g., agents can directly perceive their working environments through cameras and extract information using computer vision methods to populate the knowledge base. Note that some methods use a combination of sources, e.g., agents extract information from some Web sources provided by humans, or agents solicit information through dialog with humans. Knowledge directly encoded by domain experts is often more reliable but it may require considerable time and effort from these experts. In comparison, the human effort required to enable agents to acquire knowledge is typically much less, but this knowledge is often less reliable.

\subsection{Summary of Characteristic Factors} 
Methods that use RDK for SDM can be mapped to the space whose axes are the factors described above; these factors are also summarized in Figure~\ref{fig:choices}. Some methods can include combinations of the factors related to representation, reasoning, and/or knowledge acquisition. 
For instance, a given system could support active online knowledge acquisition while reasoning with domain dynamics and belief states, whereas another system could support interactive knowledge acquisition from humans while reasoning about actions and change based on a linked representation. 
Methods that couple representation, reasoning, and learning provide key benefits, e.g., reasoning can be used to trigger, inform, and guide efficient knowledge acquisition. However, they also present some challenges, e.g., in suitably reconciling differences between existing knowledge and learned knowledge. These advantages and challenges are discussed below in the context of representative RDK-for-SDM methods.


\section{RDK-for-SDM Methods}
\label{sec:rdkforsdm}
In this section, we review some representative RDK-for-SDM systems by grouping them based on their primary contributions. First, Section~\ref{sec:rdkforsdm-kr} discusses some systems that primarily focus on the knowledge representation challenges in RDK-for-SDM. Sections~\ref{sec:rdkforsdm-reason}-~\ref{sec:rdkforsdm-learn} then describe RDK-for-SDM systems in which the key focus is on the underlying reasoning and knowledge acquisition challenges respectively. Note that this grouping is based on \emph{our} understanding of the key contributions of each system; many of these systems include contributions across the three groups as summarized in Table~\ref{tab1}.

\subsection{Representation-focused Systems}
\label{sec:rdkforsdm-kr}
As stated in Section~\ref{sec:background-rdk}, many generic hybrid representations have been developed to support both logical and probabilistic reasoning with knowledge and uncertainty. 

\subsubsection{Unified RDK-for-SDM Representations}
Developing a unified representation for RDK and SDM maps to developing a unified representation for logical and probabilistic reasoning, which has been a fundamental problem in robotics and AI for decades. 
Frameworks and methods based on unified representations provide significant expressive power, but they also impose a significant computational burden despite the ongoing work on developing more efficient (and often approximate) reasoning algorithms for such unified paradigms. 

\smallskip
\noindent
\textbf{Statistical Relational AI} Some of the foundational work in this area has built on work in statistical relational learning/AI. These RDK-for-SDM methods typically use unified representations and  differ based on the underlying design choices. For instance, Markov Logic Networks (MLNs) combine probabilistic graphical models and first order logic, assigning weights to logic formulas~\citep{richardson:ML06}; these have been extended to Markov logic decision networks by associating logic formulas with utilities in addition to weights~\citep{nath:starai09}.
In a similar manner, Probabilistic Logic (ProbLog) programming annotates facts in logic programs with probabilities and supports efficient inference and learning using weighted Boolean formulas~\citep{deraedt:ML15}. This includes an extension of the basic ProbLog system, called Decision-Theoretic (DT)ProbLog, in which the utility of a particular choice of actions is defined as the expected reward for its execution in the presence of probabilistic effects~\citep{broeck:aaai10}. Another example of an elegant (unified) formalism for dealing with degrees of belief and their evolution in the presence of noisy sensing and acting, extends situation calculus by assigning weights to possible worlds and embedding a theory of action and sensing~\citep{bacchus:AIJ99}. This formalism has been extended to deal with decision making in the continuous domains seen in many robotics applications~\citep{belle:AIJ18}. Others have developed frameworks based on unified representations specifically for decision theoretic reasoning, e.g., first-order relational POMDPs that leverage symbolic programming for the specification of POMDPs with first-order abstractions~\citep{juba:JMLR16,sanner2010symbolic}. 

\smallskip
\noindent
\textbf{Classical Planning} RDK-for-SDM systems based on unified representations have also built on tools and methods in classical planning. Examples include PPDDL, a probabilistic extension of the action language PDDL, which retains the capabilities of PDDL and provides a semantics for planning problems as MDPs~\citep{younes2004ppddl1}, and Relational Dynamic Influence Diagram Language (RDDL) that was developed to formulate factored MDPs and POMDPs~\citep{sanner2010relational}. In comparison with PPDDL, RDDL provides better support for modeling concurrent actions and for representing rewards and uncertainty quantitatively. 

\smallskip
\noindent
\textbf{Logic Programming} RDK-for-SDM systems with a unified representation have also been built based on logic programming frameworks. One example is P-log, a probabilistic extension of ASP that encodes probabilistic facts and rules to compute probabilities of different possible worlds represented as answer sets~\citep{baral:TPLP09}. P-log has been used to specify MDPs for SDM tasks, e.g., for robot grasping~\citep{zhu2012plog}. More recent work has introduced a coherence condition that facilitates the construction of P-log programs and proofs of correctness~\citep{balai2019p}. One limitation of P-log, from the SDM perspective, is that it requires the horizon to be provided as part of the input. The use of P-log for probabilistic planning with infinite horizons requires a significant engineering effort. 



\subsubsection{Linked RDK-for-SDM Representations}

As stated earlier in the context of Factor 1 in Section~\ref{sec:factors-represent}, RDK-for-SDM systems with linked (hybrid) representations trade expressivity or correctness guarantees for computational speed, an important consideration if an agent has to respond to dynamic changes in complex domains. These methods often also use different levels of abstraction and link rather than unify the corresponding descriptions of knowledge and uncertainty. This raises interesting questions about the choice of domain variables in each representation, and the transfer of knowledge and control between the different reasoning mechanisms. For instance, a robot delivering objects in an office building may plan at an abstract level, reasoning logically with rich commonsense domain knowledge (e.g., about rooms, objects, and exogenous agents) and cognitive theories. The abstract actions can be implemented by reasoning probabilistically at a finer resolution about relevant domain variables (e.g., regions in specific rooms, parts of objects, and agent actions).


\smallskip
\noindent
\textbf{Switching Systems} The simplest option for methods based on linked representations is to switch between reasoning mechanisms based on different representations for different tasks. One example is the \textit{switching planner} that uses either a classical first-order logic planner or a probabilistic (decision-theoretic) planner for action selection~\citep{gobelbecker2011switching}. This method used a combination of the Fast-Downward~\citep{helmert2006fast} and PPDDL~\citep{younes2004ppddl1} representations. Another approach uses ASP for planning and diagnostics at a coarser level of abstraction, switches to using probabilistic algorithms for executing each abstract action, and adds statements to the ASP program's history to denote success or failure of action execution; this approach has been used for multiple robots in scenarios that mimic manufacturing in toy factories~\citep{saribatur:AR19}.  

\smallskip
\noindent
\textbf{Tightly-Coupled Systems}
There has been some work on generic RDK-for-SDM frameworks that represent and reason with knowledge and beliefs at different abstractions, and ``\emph{tightly couple}'' the different representations and reasoning mechanisms by formally establishing the links between and the attributes of the different representations. These methods are often based on the \textit{principle of refinement}~\citep{freeman1991refinement}. 
This principle has also been explored in fields such as software engineering and programming languages~\citep{lovas2010refinement,pfenning2010refinement}, but without any theories of actions and change that are important in robotics and AI. One approach examined the refinement of agent action theories represented using situation calculus at two different levels. This approach makes a strong assumption of the existence of a bisimulation relation between the action theories for a given refinement mapping between these theories at the high-level and the low-level~\citep{banihashemi:ijcai18}. The principle of refinement has also been used to construct abstractions of ASP programs, with the objective of shrinking the domain size while preserving the structure of the rules~\citep{saribatur:AIJ21}.
An example of tightly-coupled systems in robotics is the refinement-based architecture (REBA) that considers transition diagrams of any given domain at two different resolutions, with the fine-resolution diagrams defined formally as a refinement of the coarse-resolution diagram~\citep{sridharan2019reba}. Non-monotonic logical reasoning with limited commonsense domain knowledge at the coarse-resolution provides a sequence of abstract actions to achieve any given goal. Each abstract action is implemented as a sequence of concrete actions by automatically zooming to and reasoning probabilistically with automatically-constructed models (e.g., POMDPs) of the relevant part of the fine-resolution diagram, adding relevant observations and outcomes to the coarse-resolution history. The formal definition of refinement, zooming, and the connections between the transition diagrams enables smooth transfer of relevant information and control, and improves scalability. It also enables the robot to represent and reason with sophisticated cognitive theories in the coarse resolution, e.g., with an adaptive theory of intentions~\citep{gomez:AMAI21}.

\smallskip
\noindent
\textbf{Cognitive Architectures} Systems such as ACT-R~\citep{anderson2014atomic}, SOAR~\citep{laird2012soar}, ICARUS~\citep{langley:aaai06} and DIRAC~\citep{scheutz:AR07} can represent and draw inferences based on declarative knowledge, often using first-order logic. These architectures typically support SDM through a linked representation, but some architectures have pursued a unified representation for use in robotics by attaching a quantitative measure of uncertainty to logic statements~\citep{sarathy:TCDS18}.

\smallskip
\noindent
There are many other RDK-for-SDM systems based on hybrid representations. In these systems, the focus is not on developing new representations; they instead adapt or combine existing representations to support interesting reasoning and learning capabilities, as described below.

\subsection{Reasoning-focused Systems}
\label{sec:rdkforsdm-reason}
Next, we discuss some other representative RDK-for-SDM systems in which the primary focus is on addressing related reasoning challenges. 


\smallskip
\noindent

\textbf{RDK for State Estimation}
RDK methods can be used for estimating the current world state in order to guide SDM. Although practical domains often include many objects with different attributes, and multiple relationships between these objects, only a small subset of these objects, attributes, and relationships may be relevant to any particular task that an agent has to perform. Researchers have therefore used RDK methods to identify the task-relevant information to guide state estimation and SDM. For instance, the state of the world has been represented using knowledge predicates and assumptive predicates, which were then used for planning based on declarative action knowledge and probabilistic rules~\citep{hanheide2017robot}. This approach, embedded within a three-layered architecture, was used for applications such as object search, and semantic mapping.  Another example is the CORPP system that uses P-log~\citep{balai2019p} to reason with probabilistic declarative knowledge in order to generate informative priors for POMDP planning~\citep{zhang2015corpp}. In a human-robot dialog domain, CORPP demonstrated that commonsense knowledge, such as ``people like coffee in the mornings'' and ``office doors are closed over weekends'', is useful for guiding dialog actions. Other researchers have exploited factored state spaces to develop algorithms that use probabilistic declarative knowledge to efficiently compute informative priors for POMDPs~\citep{chitnis2018integrating}. In particular, they developed an efficient belief state representation that dynamically selects an appropriate factoring to guide SDM, and demonstrated its effectiveness in robot cooking tasks. These methods separate the variables modeled at different levels and (manually) link relevant variables between the levels, improving scalability and dynamic response. These links enable the flow of information between the different reasoning mechanisms, often at different abstractions, but they typically do not focus on developing (or extending) the underlying representations or on establishing the properties of the connections between the representations.


\smallskip
\noindent
\textbf{Dynamics Models for SDM}
In some RDK-for-SDM systems, the focus is on RDK guiding the construction or adaptation of the world models used for SDM.
One example is the extension of~\citep{chitnis2018integrating} that seeks to automatically determine the variables to be modeled in the different representations~\citep{chitnis2020learning}. Another example is the use of logical smoothing to refine past beliefs in light of new observations; the refined beliefs can then be used for diagnostics and to reduce the state space for planning~\citep{mombourquette2017logical}. There is also work on an action language called pBC+, which supports the definition of MDPs and POMDPs over finite and infinite horizons~\citep{wang2019bridging}. 


In some RDK-for-SDM systems, RDK and prior experiences of executing actions in the domain are used to construct domain models and guide SDM. For instance, symbolic planning has been combined with hierarchical RL to guide the agent's interactions with the world, resulting in reliable world models and SDM~\citep{illanes2020symbolic}. 
In other work, each symbolic transition is mapped (manually) to options, i.e., temporally-extended MDP actions; RDK helps compute the MDP models and policies, and the outcomes of executing the corresponding primitive actions help revise the values of state action combinations in the symbolic reasoner~\citep{yang2018peorl}. These systems use a linked representation, and reason about dynamics in RDK and states and world models in SDM. Other systems reason without explicit world models in SDM, e.g., the use of deep RL methods to compute the policies in the options corresponding to each symbolic transition in the context of game domains~\citep{lyu2019sdrl}. 


\begin{table*}[t]
\renewcommand*{\arraystretch}{1.4}
\caption {A subset of the surveyed RDK-for-SDM algorithms from the literature mapped to the space defined by the characteristic factors discussed in Section~\ref{sec:factors}. 
Each column corresponds to one characteristic factor (except for the last one); if a factor's range includes multiple values, this table shows the most typical value. 
\textbf{Uni. Rep.:} unified representation for both RDK and SDM (Factor 1). 
\textbf{Abs. Rep.:} abstract representations for RDK and SDM that are linked together (Factor 2). 
\textbf{Dyn. RDK:} declarative knowledge includes action knowledge and can be used for task planning (Factor 3). 
\textbf{RL SDM:} world models are not provided to SDM, rendering RL necessary (Factor 4). 
\textbf{Par. Obs.:} current world  states are partially observable (Factor 5). 
\textbf{On. Acq.:} online knowledge acquisition is enabled (Factor 6). 
\textbf{ML RDK:} at least part of the knowledge base is learned by the agents, where the opposite is human developing the entire knowledge base (Factor 7). 
\textbf{Rew. RDK:} RDK is used for reward shaping. 
} 
\label{tab1} 
\begin{center}
\tiny
\begin{tabular}{|l| l | c c c c c c c c|} 

\hline
  & & Uni. Rep. &  Abs. Rep.&  Dyn. RDK &  RL SDM   & Par. Obs. &  On. Acq.     & ML RDK    & Rew. RDK \\\hline 
 \parbox[t]{2mm}{\multirow{5}{*}{\rotatebox[origin=l]{90}{Representation}}} 
 &\citep{younes2004ppddl1}            & \pief     & \piee	    & \pie      & \piee     & \piee     & \piee         & \piee     & \piee	\\\cline{2-10}
 
 &\citep{sanner2010relational}        & \pief     & \piee	    & \pie      & \piee     & \pieh     & \piee         & \piee     & \pief	\\\cline{2-10}
 
 &\citep{baral:TPLP09}                & \pief     & \piee     & \pie      & \piee     & \pieh     & \piee         & \piee     & \piee \\\cline{2-10} 
 
 &\citep{wang2019bridging}            & \pief     & \piee	    & \pie      & \piee     & \pieh     & \pief         & \pief     & \pief	\\\cline{2-10} 
 
 &\citep{zhang2017dynamically}        & \pief     & \piee	    & \piee     & \piee     & \pieh     & \pief         & \piee     & \pief	\\ \hline \hline 

 \parbox[t]{2mm}{\multirow{15}{*}{\rotatebox[origin=l]{90}{Reasoning}}} 
 &\citep{sridharan2019reba}           & \piee     & \pief	    & \pief     & \piee     & \pief     & \pief         & \pief     & \piee	\\\cline{2-10} 
 &\citep{illanes2020symbolic}         & \piee     & \pief	    & \pief     & \pief     & \piee     & \piee         & \piee     & \piee \\\cline{2-10} 
 &\citep{yang2018peorl,lyu2019sdrl}   & \piee     & \pief	    & \pief     & \pief     & \piee     & \pief         & \piee     & \piee	\\\cline{2-10} 
 &\citep{furelos2020induction}        & \piee     & \pief	    & \pief     & \pief     & \piee     & \pief         & \pief     & \pief	\\\cline{2-10} 
 &\citep{gobelbecker2011switching}    & \piee     & \piee	    & \pief     & \piee     & \pief     & \piee         & \piee     & \piee	\\\cline{2-10} 
 &\citep{garnelo2016towards}
                                    & \piee     & \piee	    & \piee     & \pief     & \piee     & \pief         & \pief     & \piee	\\\cline{2-10}
 &\citep{chitnis2018integrating}
                                    & \piee     & \piee	    & \piee     & \piee     & \pief     & \pief         & \piee     & \piee	\\\cline{2-10} 
 &\citep{zhang2015mixed,zhang2015corpp} & \piee   & \piee     & \piee     & \piee     & \pief     & \piee         & \piee     & \piee	\\\cline{2-10} 
 &\citep{amiri2020learning}           & \piee     & \piee	    & \piee     & \piee     & \pief     & \pief         & \pief     & \piee	\\\cline{2-10} 
 &\citep{grounds2005combining}        & \piee     & \pief	    & \pief     & \pief     & \piee     & \piee         & \piee     & \pief	\\\cline{2-10} 
 &\citep{hoelscher2018utilizing}      & \piee     & \piee	    & \piee     & \piee     & \pief     & \pief         & \piee     & \pief	\\\cline{2-10}
 &\citep{icarte2018using,camacho2019ltl}              & \piee     & \piee	    & \pief     & \pief     & \piee     & \piee         & \piee     & \pief	\\\cline{2-10}
 &\citep{zhang2019faster}             & \piee     & \piee	    & \piee     & \pief     & \piee     & \piee         & \piee     & \piee	\\\cline{2-10}
 &\citep{leonetti2016synthesis}       & \piee     & \piee	    & \pief     & \pief     & \piee     & \pief         & \piee     & \piee	\\\cline{2-10}
 &\citep{eysenbach2019search}         & \piee     & \pief	    & \pief     & \pief     & \piee     & \pief         & \pief     & \piee	\\\hline \hline 
 \parbox[t]{2mm}{\multirow{5}{*}{\rotatebox[origin=l]{90}{Acquisition}}} 
 &\citep{konidaris2018skills,gopalan2020simultaneously}            
                                    & \piee     & \pief	    & \pief     & \piee     & \piee     & \pief         & \pief     & \piee	\\\cline{2-10} 
 
 &\citep{thomason2015learning,amiri2019augmenting}
                                    & \piee     & \piee	    & \piee     & \piee     & \pief     & \pief         & \piee     & \piee	\\\cline{2-10} 
 
 &\citep{she2017interactive}          & \piee     & \piee	    & \pief     & \pief     & \piee     & \pief         & \pief     & \piee	\\\cline{2-10} 
 
 &\citep{mericli2014interactive}      & \piee     & \piee	    & \pief     & \piee     & \piee     & \pief         & \piee     & \piee	\\\cline{2-10} 
 
 &\citep{samadi2012using}             & \piee     & \piee	    & \piee     & \piee     & \piee     & \pief         & \pief     & \piee	\\\hline

\end{tabular}
\end{center}
\end{table*}

\smallskip
\noindent
\textbf{Credit Assignment and Reward Shaping} When MDPs or POMDPs are used for SDM in complex domains, rewards are sparse and typically obtained only on task completion, e.g., after executing a plan or at the end of a board game. As a special case of learning and using world models in SDM, researchers have leveraged RDK methods to model and shape the rewards to improve the agent's decision-making. For instance, declarative action knowledge has been used to compute action sequences, using the action sequences to compute a potential function and for reward shaping in game domains~\citep{grounds2005combining,grzes2008plan,efthymiadis2013using}. In this work, RL methods such as Q-learning, SARSA, and Dyna-Q were combined with a STRIPS planner, with the planner shaping the reward function used by the agents to compute the optimal policy. These systems perform RDK with domain dynamics, and reason about states but no explicit world models in SDM. 

In some cases, the reward specification is obtained from statistics and/or contextual knowledge provided by humans.
For example, the iCORPP algorithm enables a robot to reason with contextual knowledge using P-log to automatically determine the rewards (and transition functions) of a POMDP used for planning~\citep{zhang2017dynamically}. 
Another system, called LPPGI, enables robots to leverage human expertise for POMDP-based planning under uncertainty in the context of task specification and execution~\citep{hoelscher2018utilizing}. RDK in this system is rather limited; domain dynamics are not considered and the system is limited to maximizing the expected probability of satisfying logic objectives in the context fo a robot arm stacking boxes. 
There has also been work on ``reward machines'' that uses Linear Temporal Logic to represent and reason with declarative knowledge, especially temporal constraints implied by phrases such as ``until'' and ``eventually,'' 
in order to automatically generate additional rewards for RL that are potentially non-Markovian~\citep{icarte2022reward}. 


\smallskip
\noindent
\textbf{Guiding SDM-based Exploration} 
When the main objective of SDM is exploration or discovery of particular aspects of the domain, RDK can be used to inform and guide the trade-off between exploration and exploitation, and to avoid poor-quality exploration behaviors in SDM. For instance, the DARLING algorithm uses RL to explore and compute action sequences that lead to long-term goals under uncertainty, with RDK being used to filter out unreasonable actions from exploration~\citep{leonetti2016synthesis}; this approach has been evaluated on real robots navigating office environments to locate people of interest. 

An algorithm called GDQ uses action knowledge to generate artificial, ``oppotimistic'' experience to give RL agents a warm-up learning experience before letting them interact with the real world~\citep{hayamizu2021guiding}. 
Another similar approach uses RDK to guide an agent's exploration behavior (formulated as SDM) in non-stationary environments~\citep{ferreira2017answer}, and to learn constraints that prevent risky behaviors in video games~\citep{zhang2019faster}. There is also work on non-monotonic logical reasoning with commonsense knowledge to automatically determine the state space for relational RL-based exploration of previously unknown action capabilities~\citep{mohan:icaps17}.

\subsection{Knowledge Acquisition-focused Systems}
\label{sec:rdkforsdm-learn}
Next, we discuss some RDK-for-SDM systems whose main contribution is the acquisition (and revision) of domain knowledge used for RDK. This knowledge can be obtained through manual encoding and/or automated acquisition from different sources (Web, corpora, sensor inputs). 

\smallskip
\noindent
\textbf{Knowledge Acquisition while Acting} Some RDK-for-SDM systems allow the agent to acquire knowledge while also simultaneously reasoning and executing actions in dynamic domains. Such systems can often support online and offline knowledge acquisition, with active and reactive aspects. For example, ASP-based non-monotonic logical reasoning has been used to guide relational RL (i.e., SDM) and decision-tree induction in order to learn previously unknown actions and domain axioms; this knowledge is subsequently used for RDK~\citep{mohan:ACS18}. This system supports reactive knowledge acquisition, with reasoning used to trigger and guide learning only when some unexpected outcomes are observed (e.g., to acquire knowledge of previously unknown constraints), as well as active, online knowledge acquisition, with the robot acquiring previously unknown knowledge based on explicit exploration (e.g., of the potential effects of new actions).

\smallskip
\noindent
\textbf{Knowledge Acquisition from Experience} There is a well established literature of RDK-for-SDM systems, including many described above, acquiring or revising knowledge of domain dynamics in a  supervised or semi-supervised \emph{training} phase. The robot could, for instance, be asked to execute different actions and observe the corresponding outcomes in scenarios with known ground truth information~\citep{sridharan2019reba,zhang2017dynamically}. More recently, some RDK-for-SDM systems have built on recent developments in data-driven methods (e.g., deep learning and RL) to acquire knowledge. For instance, the symbols needed for task planning have been extracted from the replay buffers of multiple trials of deep RL, with similar states (in the replay buffers) being grouped to form the search space for symbolic planning~\citep{eysenbach2019search}. In robotics domains, a small number of real-world trials have been used to enable a robot to learn the symbolic representations of the preconditions and effects of a door-opening action~\citep{konidaris2018skills}. Knowledge acquisition in these systems is often offline (i.e., batch of data collected from the robot is processed offline to extract knowledge); this acquisition can be achieved by targeted exploration (i.e., active) or reactive. Researchers have also enabled robots to simultaneously acquire latent space symbols and language groundings based on prior demonstration trajectories paired with natural language instructions~\citep{gopalan2020simultaneously}; in this case, knowledge acquisition is active and offline, and requires significantly fewer training samples compared to end-to-end systems. In another RDK-for-SDM system, non-monotonic logical reasoning is used to guide deep network learning and active acquisition of previously unknown axioms describing the behavior of these networks~\citep{mota:SNCS21,riley:Frontiers19}.

\smallskip
\noindent
\textbf{Knowledge Acquisition from Humans, Web, and other sources} For some RDK-for-SDM systems, researchers have developed a dialog-based interactive approach for situated task specification, with the robot learning new actions and their preconditions through verbal instructions~\citep{mericli2014interactive}. In a related approach, SDM has been used to manage human-robot dialog, which helps a robot acquire knowledge of synonyms (e.g., ``java" and ``coffee") that are used for RDK~\citep{thomason2015learning}. Building on this work, other researchers have developed methods to add new object entities to the declarative knowledge in RDK-for-SDM systems~\citep{amiri2019augmenting}. In other work, human (verbal) descriptions of observed robot behavior have been used to extract knowledge of previously unknown actions and action effects, which is merged with existing knowledge in the RDK component~\citep{mohan:ACS18}. More recent work in the context of a system enabling an agent to respond to a human's questions about its decisions and evolution of beliefs, has also enabled the agent to interactively construct questions to resolve ambiguities in the human's questions~\citep{mota:icdl21}.

Some researchers have equipped their RDK-for-SDM systems with the ability to acquire domain knowledge using data available on the Web~\citep{samadi2012using}. Information (to be encoded in first-order logic) about the likely location of paper would, for instance, be found by analyzing the results of a web search for ``kitchen" and ``office".
\section{Challenges and Opportunities}
\label{sec:challenges-opportunities}
Over the last few decades, researchers have made significant progress in developing sophisticated methods for reasoning with declarative knowledge and for sequential decision making under uncertainty. In recent years, improved understanding of the complementary strengths of the methods developed in these two areas has also led to the development of sophisticated methods that seek to integrate and exploit these strengths. These integrated systems have provided promising results, but they have also identified several open problems and opened up many directions for further research. Below, we discuss some of these problems and research directions:
    \paragraph{Representational Choices:}
    As discussed in Section~\ref{sec:rdkforsdm-kr}, existing methods integrating RDK and SDM methods are predominantly based on unified or linked representations. General-purpose methods often use a unified representation and associated reasoning methods for different descriptions of domain knowledge, e.g., a unified representation for logic-based and probabilistic descriptions of knowledge. On the other hand, integrated systems developed specifically for robotics and other dynamic domains link rather than unify the different representations, including those at different abstractions, trading correctness for computational efficiency. A wide range of representations and reasoning methods are possible within each of these two classes; these need to be explored further to better understand the choice (of representation and reasoning methods) best suited to any particular application domain. During this exploration, it will be important to carefully study any trade-offs made in terms of the expressiveness of the representation, the ability to support different abstractions, the computational complexity of the reasoning methods, and the ability to establish that the behavior of the robot (or agent) equipped with the resulting system satisfies certain desirable properties. These hybrid representations can also form the foundation of modern neuro-symbolic AI~\citep{hitzler2022neuro,garcez2019neural} methods for reasoning and learning.
    
    \paragraph{Interactive Learning:}
    Irrespective of the representation and reasoning methods used for RDK, SDM, or a combination of the two, the knowledge encoded will be incomplete and/or cease to be relevant over a period of time in any practical, dynamic domain. In the age of ``big data", certain domains provide ready availability of a lot of labeled data from which the previously unknown information can be learned, whereas such labeled training data is scarce in other domains; in either case, the knowledge acquired from the data may not be comprehensive. Also, it is computationally expensive to learn information from large amounts of data. Incremental and interactive learning thus continues to be an open problem in systems that integrate RDK and SDM. Promising results have been obtained by methods that promote efficient learning by using reasoning to trigger learning only when it is needed and limit (or guide) learning to those concepts that are relevant to the tasks at hand (see discussion in Sections~\ref{sec:rdkforsdm-reason} and~\ref{sec:rdkforsdm-learn}); such methods need to be developed and analyzed further. Another interesting research thrust is to learn \emph{cumulatively} from the available data and merge the learned information with the existing knowledge such that reasoning continues to be efficient as additional knowledge is acquired over time~\citep{laird:IS17,langley:aaai17}. 
    
    \paragraph{Human ``in the loop'':}
    Many methods for RDK, SDM, or RDK-for-SDM, assume that any prior knowledge about the domain and the associated tasks is provided by the human in the initial stages, or that humans are available during task execution for reliable feedback and supervision. These assumptions do not always hold true in practice.
    Humans can be a rich source of information but there is often a non-trivial cost associated with acquiring and encoding such knowledge from people. Since it is challenging for humans to accurately specify or encode domain knowledge in complex domains, there is a need for methods that consider humans as collaborators to be consulted by a robot based on necessity and availability. 
    Such methods will need to address key challenges related to the protocols for communication between a robot and a human, considering factors such as the expertise of the human participants and the availability of humans in social contexts~\citep{rosenthal2012someone}. 
    Another related problem that is increasingly getting a lot of attention is to enable a reasoning and learning system to \emph{explain} its decisions and beliefs in human-understandable terms.  
    
    \paragraph{Combining Reasoning, Learning, and Control:}
    As discussed in this paper, many methods that integrate RDK and SDM focus on decision making (or reasoning) tasks. There are also some methods that include a learning component and some that focus on robot control and manipulation tasks. However, robots that sense and interact with the real world often require a system that combines reasoning, learning, and control capabilities~\citep{garrett2021integrated}. Similar to the combination of reasoning and learning (as mentioned above), tightly coupling reasoning, learning, and control presents unique advantages and unique open problems in the context of integrated RDK and SDM. 
    For instance, reasoning with predictive models and learning can be used to identify (on demand) and revise the relevant variables in the control laws for the tasks at hand~\citep{mathew:ichr19,sidhik:iros21}. At the same time, real world control tasks often require a very different representation of domain attributes, e.g., reasoning to move a manipulator arm may be performed in a discrete, coarser-granularity space of states and actions whereas the actual manipulation tasks being reasoned about need to be performed in a continuous, finer-granularity space. There is thus a need for systems that integrate RDK and SDM, and suitably combine reasoning, learning, and control by carefully exploring the effect of different representational choices and the methods being used for reasoning and learning. 

    \paragraph{Scalability and Teamwork:}
    Despite considerable research, algorithms for RDK, SDM, or a combination of the two, find it difficult to scale to more complex domains. This is usually due to the space of possible options to be considered, e.g., the size of the data to be reasoned with by the RDK methods, and the size of the state-action space to be considered by the SDM methods. All of these challenges are complicated further when applications require a team of robots and humans to collaborate with each other. For instance, representational choices and reasoning algorithms may now need to carefully consider the capabilities of the teammates before making a decision. As described earlier, there are some promising avenues to be explored further. 
    These include the computational modeling and use of principles such as relevance, persistence, and non-procrastination, which are well-known in cognitive systems~\citep{langley:aaai17}, in the design of the desired integrated system~\citep{blount:iclp15,gomez:AMAI21}. Such a system could then automatically determine the best use of available resources and algorithms depending on the domain attributes and tasks at hand.
    
    \paragraph{Explainability and Trust:} With the increasing use of AI and machine learning methods in different applications, there is renewed focus within the research community on enabling humans to understand the operation of these methods~\citep{anjomshoae:aamas19,miller:AIJ19}. Issues such as explainability  or trust remain open problems for RDK-for-SDM systems, especially those that integrate reasoning and learning in complex domains. At the same time, the design of these systems provides promising research threads to be explored further. For instance, the use of logics for representing and reasoning with commonsense knowledge in the RDK component of such systems provides a foundation for making the associated reasoning and learning more transparent. Research also indicates that the underlying representation and established knowledge representation tools can be exploited to reliably and efficiently trace beliefs and provide on-demand explanations at the desired level of abstraction, before, during, or after task execution~\citep{mohan:KI19,mota:SNCS21}. A key challenge would be rigorously study trust and explainability from the viewpoint of a non-expert human interacting with these systems. 

    \paragraph{Evaluation Measures and Benchmarks:}
    The complexity of the components of RDK-for-SDM systems, and the connections between of these components, make it rather challenging to isolate and evaluate the impact of the underlying representation, reasoning methods, and learning methods. Often, the observed performance of a particular algorithm (e.g., for planning) is influenced by the design of this algorithm and the connections between this algorithm and other methods in the system. A key direction for further research is the definition of common measures and tasks for the evaluation of such architectures; doing so would provide deeper insights into the development and use of such architectures. The evaluation measures will need to go beyond measuring the accuracy and computational efficiency of individual components (e.g., planning and task completion accuracy, learning rate, execution time) to examine the effects of the links between the components. These measures could, for instance, explore scalability to more complex domains and tasks. Here, complexity could refer to the type and amount of knowledge encoded in the system; the type, duration, and number of operations to be performed by the robot; and the number and duration of interactions between the different components (of the system) required to complete the task. In addition, evaluation could consider qualitative measures of performance, e.g., the ability to complete different tasks, the ability to provide interactive explanations, or the satisfaction of humans interacting with the system. 
    
    The benchmarks used for evaluation should not be limited to providing datasets or scenarios for evaluating individual algorithms. Similar to the evaluation measures, the benchmarks should instead challenge the robot to explore and use the interplay between the different components of the system being evaluated, e.g., use reasoning to guide knowledge acquisition, and use the learned knowledge to inform reasoning. In this context, many different domains hold promise in terms of being suitable for evaluation of such RDK-for-SDM systems; these include \textit{games}~\citep{yang2018peorl,zhang2019faster}, \textit{interactive dialog}~\citep{zhang2015corpp,amiri2019augmenting}, \textit{robot navigation and exploration}~\citep{leonetti2016synthesis,hanheide2017robot}, and \textit{scene understanding}~\citep{chitnis2018integrating,mota:rss19,jiang2019open,mota:SNCS21}.


\section*{Acknowledgments}
\noindent
Related work in the Autonomous Intelligent Robotics (AIR) group at SUNY Binghamton was supported in part by grants from NSF (NRI-1925044), Ford Motor Company (URP Awards), OPPO (Faculty Research Award), and SUNY RF. Related work in the Intelligent Robotics Lab (IRLab) at the University of Birmingham was supported in part by the U.S. Office of Naval Research Science of Autonomy Awards N00014-17-1-2434 and N00014-20-1-2390, the Asian Office of Aerospace Research and Development award FA2386-16-1-4071, and the UK Engineering and Physical Sciences Research Council award EP/S032487/1. The authors thank collaborators on research projects that led to the development of the ideas described in this paper.

{
\bibliographystyle{named}
\bibliography{ref}
}

\section*{Autobiographical Sketch and Photograph}

\begin{wrapfigure}{rb}{.12\textwidth}
\vspace{-.5em}
\begin{center}
  \includegraphics[width=.12\textwidth]{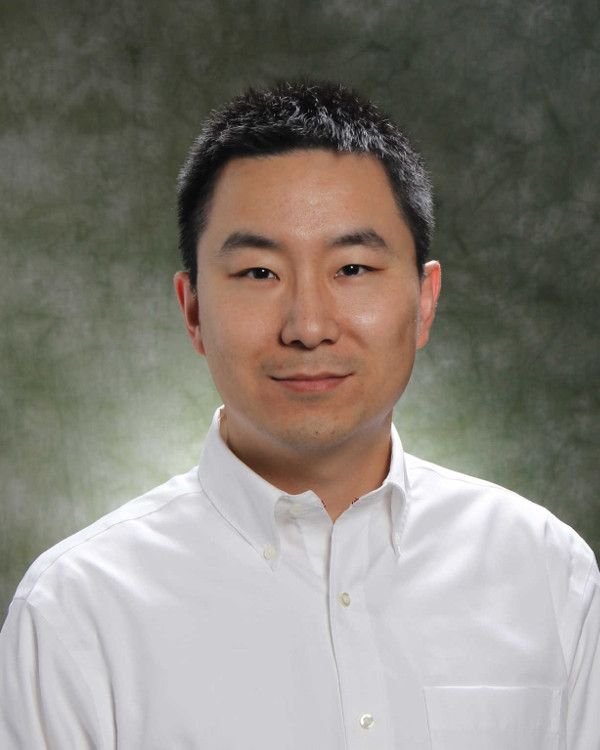}
\end{center}
\vspace{-1em}
\end{wrapfigure}
\textbf{Dr.~Shiqi Zhang} is an Assistant Professor of Computer Science, at the State University of New York (SUNY) at Binghamton (USA). 
He was a Postdoctoral Fellow at The University of Texas at Austin (USA) from 2014 to 2016, and  received his Ph.D. in Computer Science (2013) from Texas Tech University (USA). Before that, he received his Master's and B.S. from Harbin Institute of Technology in China. Dr. Zhang's research lies at the intersection of artificial intelligence and robotics. 

\vspace{1em}
\begin{wrapfigure}{rb}{.12\textwidth}
\vspace{-1.5em}
\begin{center}
  \includegraphics[width=.12\textwidth]{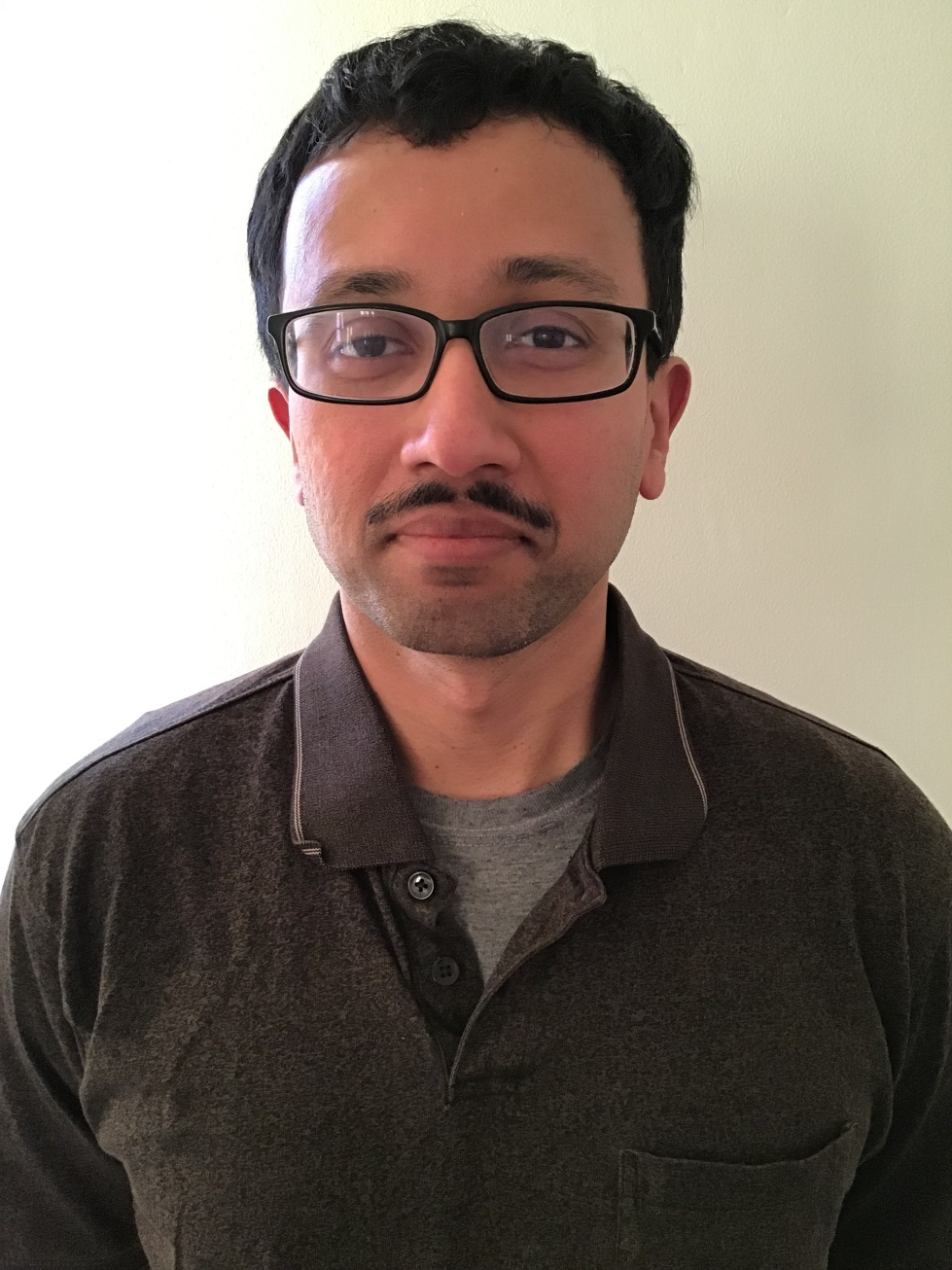}
\end{center}
\vspace{-1em}
\end{wrapfigure}
\noindent
\textbf{Dr. Mohan Sridharan} is a Reader in Cognitive Robot Systems in the School of Computer Science at the University of Birmingham (UK). Prior to his current appointment, he held faculty positions at Texas Tech University (USA) and The University of Auckland (NZ). He received his Ph.D. in Electrical and Computer Engineering from The University of Texas at Austin (USA). Dr.~Sridharan's research interests include cognitive systems, knowledge representation and reasoning, machine learning, and computational vision in the context of human-robot and human-agent collaboration.

\end{document}